\pdfoutput=1

\documentclass[11pt]{article}

\usepackage{ACL2023}

\usepackage{times}
\usepackage{latexsym}

\usepackage[T1]{fontenc}

\usepackage[utf8]{inputenc}

\usepackage{microtype}

\usepackage{inconsolata}

\usepackage{enumitem}
\usepackage{color} 
\usepackage{graphicx}
\usepackage{stfloats}
\usepackage{booktabs}
\usepackage{subcaption}
\usepackage{todonotes}
\usepackage{array}
\usepackage{amssymb, amsmath}
\usepackage{multicol}
\usepackage{multirow}
\usepackage{threeparttable}
\usepackage{algorithm}
\usepackage{algorithmic}
\usepackage{microtype}
\usepackage{lipsum,graphicx}

\usepackage{soul}
\usepackage{xcolor}
\usepackage{makecell}
\usepackage{caption}

\title{CoAD: Automatic Diagnosis through Symptom and Disease Collaborative Generation}
\author{
  Huimin Wang$^{1*}$, Wai-Chung Kwan$^{2,3}$\thanks{\ \ Equal Contribution}, Kam-Fai Wong$^{2,3}$, Yefeng Zheng$^{1}$ \\
  $^1$Jarvis Lab, Tencent, Shenzhen, China \\
  $^2$The Chinese University of Hong Kong, Hong Kong, China \\
  $^3$MoE Key Laboratory of High Confidence Software Technologies, China \\
    \texttt{\{hmmmwang,yefengzheng\}@tencent.com} \\
    \texttt{\{wckwan,kfwong\}@se.cuhk.edu.hk}
 }

\begin{document}
\maketitle
\begin{abstract}
Automatic diagnosis (AD), a critical application of AI in healthcare, employs machine learning techniques to assist doctors in gathering patient symptom information for precise disease diagnosis. The Transformer-based method utilizes an input symptom sequence, predicts itself through auto-regression, and employs the hidden state of the final symptom to determine the disease. Despite its simplicity and superior performance demonstrated, a decline in disease diagnosis accuracy is observed caused by 1) a mismatch between symptoms observed during training and generation, and 2) the effect of different symptom orders on disease prediction. To address the above obstacles, we introduce the CoAD, a novel disease and symptom collaborative generation framework, which incorporates several key innovations to improve AD: 1) aligning sentence-level disease labels with multiple possible symptom inquiry steps to bridge the gap between training and generation; 2) expanding symptom labels for each sub-sequence of symptoms to enhance annotation and eliminate the effect of symptom order; 3) developing a repeated symptom input schema to effectively and efficiently learn the expanded disease and symptom labels. We evaluate the CoAD framework using four datasets, including three public and one private, and demonstrate that it achieves an average 2.3\% improvement over previous state-of-the-art results in automatic disease diagnosis. For reproducibility, we release the code and data at \href{https://github.com/KwanWaiChung/coad}{https://github.com/KwanWaiChung/coad}. 

\end{abstract}

\section{Introduction}
The healthcare industry worldwide is facing an acute shortage of healthcare professionals such as doctors, nurses, and other staff, which results in millions of people not receiving the care they need, particularly in low-income countries \cite{world2016health}. Artificial intelligence (AI) has the potential to revolutionize medicine by automating tasks traditionally done by humans, reducing the time and cost of such tasks. Automatic diagnosis (AD) is a valuable application of AI in healthcare that aims to improve patient outcomes. When deployed on mobile devices, the AD agent functions as a chatbot, querying patients about their symptoms and health concerns, and directing them to the appropriate care based on the diagnosis. This allows for faster treatment decisions, prompt notification of care teams, and increased communication between providers, ultimately leading to improved patient outcomes and potentially saving lives. The process of AD can be conceptualized as a series of questions and answers. 
\begin{figure}[t]
\centering
\includegraphics[width=0.7\columnwidth]{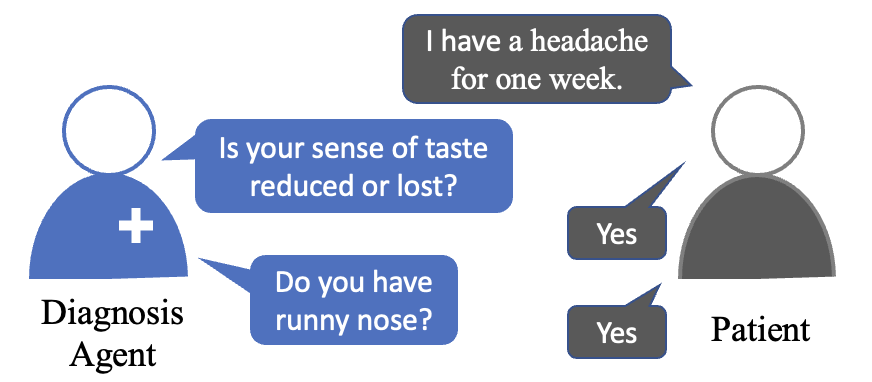} 
\caption{An example of automatic diagnosis procedure.}
\label{fig_example}
\end{figure}
As depicted in Figure \ref{fig_example}, the diagnosis begins with the patient reporting initial symptoms (in this example, only a headache). The AD agent then determines whether to ask for additional symptoms or provide a disease diagnosis. The agent carefully selects relevant queries to gather more information about the patient's condition. In this illustration, the agent chose to inquire about two specific symptoms, “reduced or lost sense of taste” and “runny nose”, and received positive responses from the patient. The AD agent has two objectives during the question-and-answer process for diagnosis. Firstly, it strategically selects symptoms that provide the most information for disease discrimination. Secondly, it aims to identify the disease as accurately as possible. These goals are interdependent. If the agent fails to thoroughly gather information about the patient's symptoms, it risks missing crucial information related to the underlying disease, resulting in an incorrect diagnosis. On the other hand, an accurate disease diagnosis enables the agent to cluster symptoms associated with the disease, thereby improving its ability to ask relevant symptom inquiries.

A significant group of existing methods approaches AD as a sequential decision-making problem and employs reinforcement learning (RL) to solve it \cite{kao2018context, peng2018refuel, xu2019end, 10.1093/bioinformatics/btac408}. These methods focus on enhancing the RL agent's performance in disease diagnosis by incorporating medical knowledge, fine-tuning reward functions, and utilizing hierarchical model structures. However, they often overlook the agent's ability to gather symptoms. In reality, it is challenging for an RL agent to simultaneously optimize for both accurate symptom inquiry and disease diagnosis. We observed in our experiments that there were fluctuations in the performance of diseases and symptoms and the RL agent frequently fell into local minima. Additionally, many RL methods use an immediate reward of “-1” to encourage shorter turns during training, which may be beneficial for other task-oriented dialogs such as ticket-booking, but is unhelpful for training the agent in symptom inquiry. As a result, most RL-based agents only ask for one or two symptoms before prematurely moving on to disease prediction. Insufficient symptom information not only leads to incorrect disease judgment but also diminishes the reliability and interpretability of the decision. 
To address the limitations of RL-based methods, \citet{chen2022diaformer} proposed a generation model to acquire symptom inquiry logic and three training tasks to train the AD agent to master symptom generation rules. Their method achieved competitive results, particularly in symptom recall, which confirms the superiority of generation-based models for diagnosis modeling. However, they did not consider the benefits of disease information. 



In this paper, we propose CoAD, a Transformer-decoder-based framework for Automatic Diagnosis that collaboratively generates disease and symptom information. CoAD utilizes three key techniques to enhance performance. First, it aligns disease labels with symptom inquiry steps, bridging the gap between training and generation for improved disease diagnosis. Second, it expands symptom labels to each sub-sequence with subsequent symptoms, making training more robust to varying symptom orders. Lastly, it incorporates a repeated symptom input schema and an additional symptom attention module for simultaneous optimization. 

Our main contributions include:

\begin{itemize}    
    \item A novel framework that effectively aligns disease labels with symptom steps, bridging the gap between training and generation. 

   \item A symptom label augmentation mechanism that strengthens training signals and enhances disease diagnosis, regardless of symptom order.
    
    \item An approach that combines repeated symptom input and symptom attention schema for concurrent symptom and disease generation.
\end{itemize}



    

\section{Related Work}
\paragraph{\textbf{RL-based approaches.}} Early work in automatic diagnosis often used the Markov decision process framework and employed reinforcement learning (RL) to solve the problem. For example, \citet{wei2018task} approached it as a task-oriented dialog task and trained an agent using deep Q-learning to collect implicit symptoms during patient interactions. To improve the poor efficiency of RL-based methods, \citet{kao2018context} added contextual information to the dialog state and applied a hierarchical RL schema by dividing diseases into different body parts. Similarly, \citet{10.1093/bioinformatics/btac408} employed a hierarchical setting, using a master model to trigger a low-level model comprised of symptom checkers and a disease classifier.
\citet{peng2018refuel} developed an RL-based diagnosis model that incorporated reward shaping to guide the search for better policies and feature rebuilding to improve the agent's ability to learn correlations between features. They also integrated domain knowledge to make more accurate decisions. Similarly, \citet{xu2019end} demonstrated that incorporating medical knowledge through a knowledge-routed graph improved diagnosis performance. Furthermore, \citet{xia2020generative} applied a GAN-based approach to the diagnosis problem, using the generator as a policy network, the discriminator as part of the reward function, and mutual information to encourage the model to select the most discriminative symptoms for diagnosis.
\paragraph{\textbf{Generation-based approaches.}}
Generation-based diagnosis methods have shown promise in their ability to predict symptoms with stronger performance compared to reinforcement learning (RL)-based methods. For example, \citet{lin2019enhancing} used a Bi-LSTM to encode symptom word sequences and trained a sequence-to-sequence model with a CRF decoder to recognize symptoms based on a symptom graph, document-level and corpus-level attentions. More recently, \citet{chen2022diaformer} aimed to alleviate the inefficiencies of exploration and sparse rewards in RL by formulating the diagnosis process as a sequence generation problem and using a Transformer-based network to learn symptom sequences and disease classifiers through three training tasks. Their model demonstrated significant improvements over RL-based counterparts.

\section{Sequence Generation based AD}
\label{sec: adsg}
The process of AD involves predicting a disease by asking a series of questions about potential symptoms from a patient who has provided initial symptoms \cite{peng2018refuel}. AD aims to optimize for two goals simultaneously: 1) asking questions to gain the most information about the patient's condition, and 2) identifying the disease quickly and accurately. In this study, we approach the problem of AD as a combined task of generating a sequence of symptoms and classifying the disease.
Let $\mathcal{S}$ denote the set of symptoms and $\mathcal{D}$ denote the set of diseases. An AD dataset considers a set of possible symptom profiles: $X_s = {s^1_E, \cdots , s^N_E, s^1_I,\cdots, s^M_I}$ with the symptoms' status of True, False, Uncertain, where $s^i_E, (i = 1, 2, \cdots, N)$ are the initial reported symptoms (i.e., explicit symptoms), and $N$ is the number of explicit symptoms; $s^i_I, (i = 1, 2, \cdots , M)$ are the subsequent acquired symptoms (i.e., implicit symptoms), and $M$ is the number of implicit symptoms. In the proposed method, CoAD, we represent a symptom status as 1 if the patient confirms having that symptom, 2 if the patient does not have it, or 0 if the patient is uncertain about having it or not. During diagnosis, CoAD will inquire about a symptom $s \in \mathcal{S}$ or produce an end token, signaling the end of symptom checking and switching to predicting a disease $d \in \mathcal{D}$.

During the diagnosis process, the AD system that interacts with a patient can be thought of as an agent. The agent's goal is to ask questions about key symptoms that will lead to a quick and accurate disease diagnosis. Since symptoms are acquired in chronological order, it is common to factorize the joint probabilities of symptoms as the product of conditional probabilities:
\begin{equation}
\begin{aligned}
    \mathcal{P}_{\theta}(s^{1:M}_I|s^{1:N}_E) = 
    \prod_{K = 1}^{M}
    \mathcal{P}_{\theta}(s^K_I|s^{1:K-1}_I, s^{1:N}_E),
    \label{e:1}
\end{aligned}
\end{equation}
where $s^{1:N}_E$ denotes $\{s^1_E, \cdots, s^N_E\}$ and $s^{1:M}_I$ denotes $\{s^1_I, \cdots, s^M_I\}$. In practice, the probabilities can be approximated by a network parameterized by $\theta$. Once enough symptoms that aid in distinguishing the disease has been acquired, the agent proceeds to make a disease prediction. Let $d^*$ denote the given disease label of the symptoms. The agent's goal is to learn a strategy that can select $d^*$ with a high probability $ \mathcal{P}_{\theta}(d^*|s^{1:M}_I, s_E^{1:N})$. 


\paragraph{\textbf{Disease Accuracy, Symptom Recall, and Combined Score.}}


Enhancing symptom recall may lead to a decrease in disease accuracy, meaning that the agent's diagnostic performance could suffer when it achieves optimal symptom recall. This is because most previous models with higher symptom recall tend to have longer turns in a limited-turns setting. Longer sequences are less common in the training set, resulting in a distribution mismatch between training and testing that hinders the model's ability to accurately identify diseases. Consequently, disease accuracy and symptom recall tend to have an inverse correlation during testing in limited-turns settings. To comprehensively measure the diagnosis performance, we introduce a combined score, $Cs = \frac{2\cdot Rc \cdot Ac }{(Rc + Ac)}$, which indicates the tradeoff between high disease-prediction accuracy and high symptom-acquisition recall. 

\section{Symptom and Disease Collaborative Generation Framework}
\label{sec-co-gen}
\begin{figure*}[t]
\centering
\includegraphics[width=0.92\textwidth]{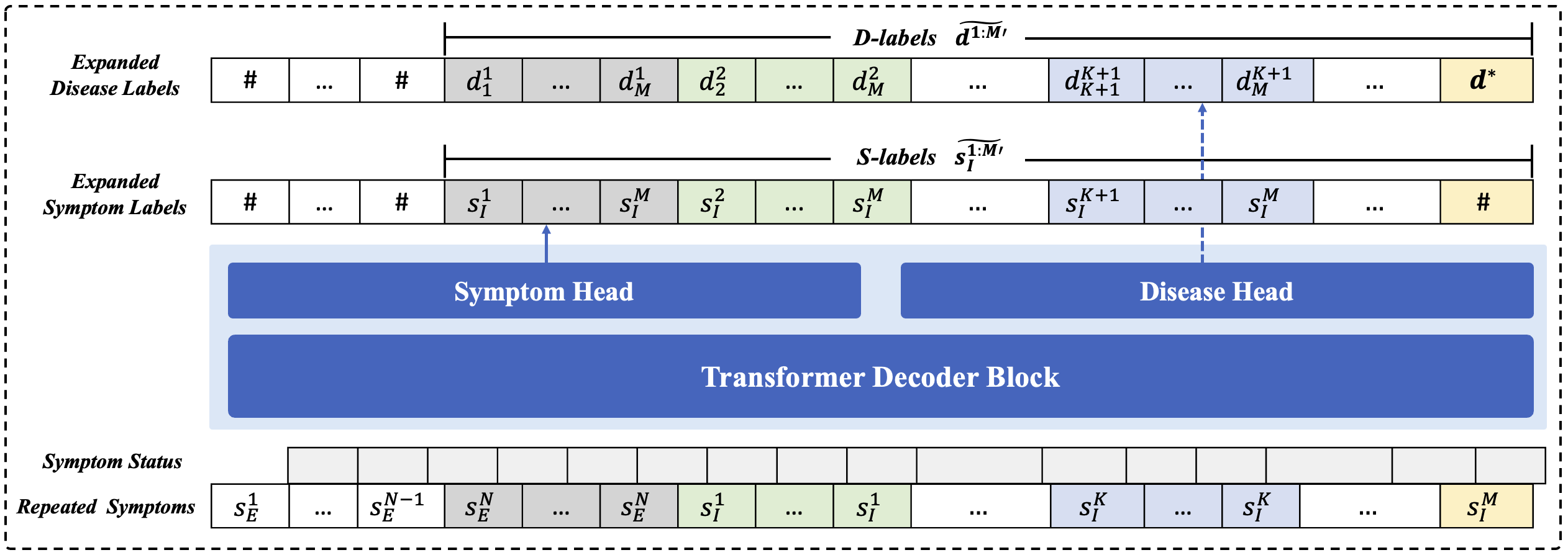}
\caption{illustration of CoAD framework, where the input explicit symptoms are $s_E^1, \cdots, s_E^N$, the implicit symptoms are $s_I^1, \cdots, s_I^M$, and the disease label is $d^*$. It takes the repeated symptom input along with the symptom type and the symptom mask as input and aims to predict the $d$-labels and $s$-labels to learn the diagnostic logic. 
}
\label{fig_method}
\end{figure*}

Even though the generation-based method has great potential in AD \cite{chen2022diaformer}, applying this method to the disease diagnosis task faces two challenges. First, the disease classifier is trained on a complete symptom sequence but is only tested on partial symptoms, which creates a gap between the visible symptoms in training and inference. Second, the order of symptoms in the training set can be inconsistent with the generated ones, which can lead to a wrong diagnosis if the symptoms in different order are incorrectly identified as different symptom sequences. To address the first challenge, we propose a disease and symptom collaborative generation framework, in which the disease label $d^*$ is expanded to multiple possible symptom steps to fill the gap of supported symptoms between training and generation. We refer to this disease label expansion procedure as \textbf{\textit{d}-label alignment}. To tackle the second challenge, we present a symptom label augmentation mechanism (called \textbf{\textit{s}-label augmentation}) that enriches the annotation and eliminates the impact of the symptoms order. Furthermore, we design a repeated symptom input schema for efficient and effective learning from the expanded disease and symptom labels. Our strategies are illustrated in Figure \ref{fig_method}, in which a Transformer decoder takes the repeated symptom embedding, symptom status embedding, and symptom mask as input. Additionally, a symptom head and a disease head are equipped to predict the $s$-labels and $d$-labels, respectively. We will provide more details in the following sections.

\subsection{$d$-label Alignment} The idea behind $d$-label alignment is straightforward: we assign the disease label $d^*$ of a symptom sequence $\{s^{1:N}_E, s^{1:M}_I\}$ to each available implicit symptom $s^K_I \in s^{1:M}_I$. A symptom $s$ is considered available if the sub-sequence $\{s^{1:N}_E, s^{1:K}_I\}$ is not present in the training set. The number of generated symptoms is much less than the number of symptoms that a patient actually has. The key to making this work is through data augmentation, transforming one symptom sequence with one disease label into multiple symptom sequences corresponding to the same disease label. This augmentation helps to fill the gap in symptoms supporting diagnoses between training and generation. At the same time, filtering out the unavailable symptoms reduces the impact of $d$-label alignment on the samples with disease labels. Additionally, $d$-label alignment results in a disease generation task, since each symptom corresponds to a disease label if we assign a special token to unavailable symptoms that are ignored during training. This enables the joint probabilities of disease sequences over symptoms to be factored as the product of conditional probabilities:

\begin{equation}
\setlength{\abovedisplayskip}{3pt}
\setlength{\belowdisplayskip}{3pt}
\begin{aligned}
    \mathcal{P}_{\theta}(d^*, d^{1:M}& |s^{1:M}_I, s^{1:N}_E) = \mathcal{P}_{\theta}(d^*|s^{1:M}_I, s^{1:N}_E) \\ &\prod_{K = 1}^{M} \mathcal{P}_{\theta}(d^K|s^{1:K-1}_I, s^{1:N}_E),
    \label{e:2}
\end{aligned}
\end{equation}
where $d^{1:M}$ are the assigned disease labels to the implicit symptoms, and $d^*$ is the originally given label for the piece of the sample. 

\begin{figure}[htbp]
\centering
\includegraphics[width=0.98\columnwidth]{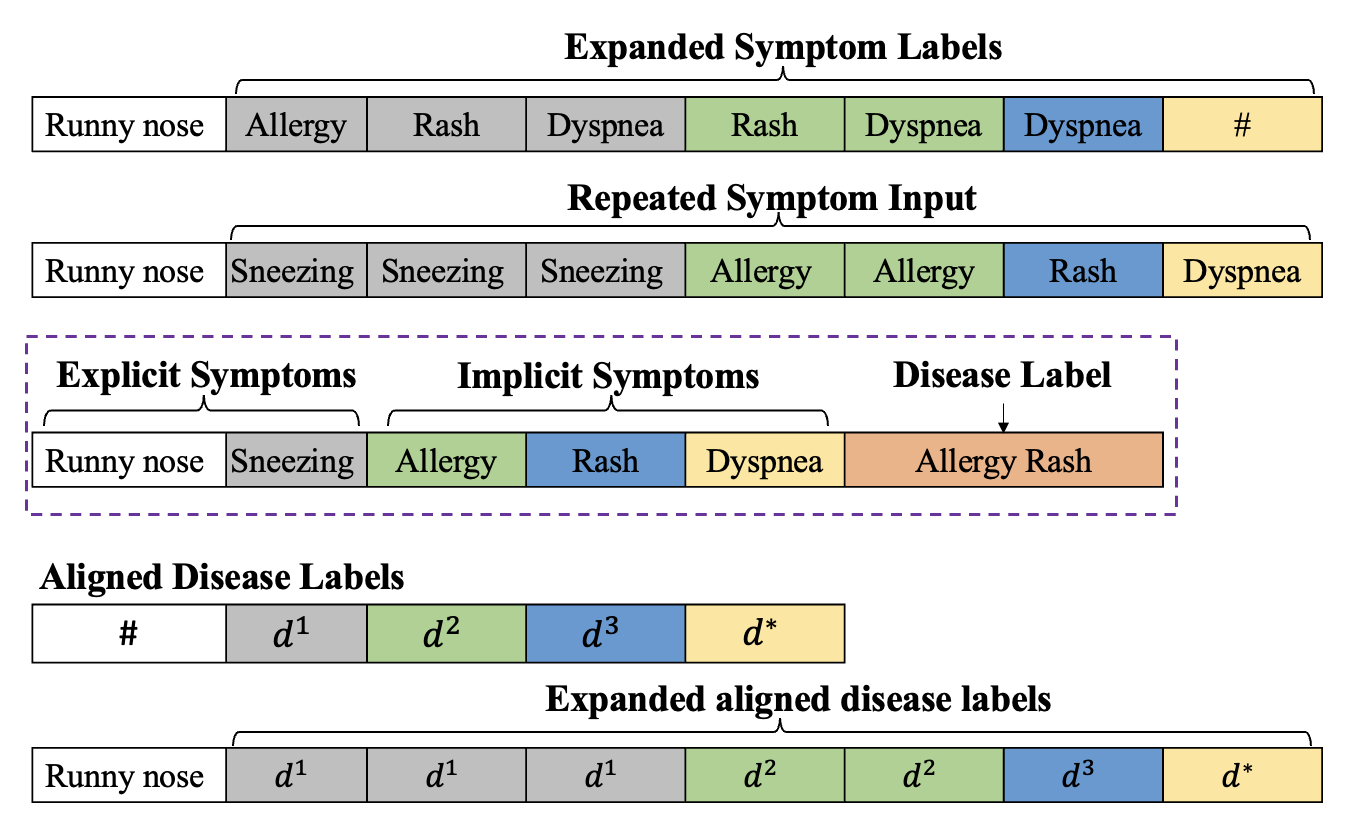} 
\caption{An example of how symptom labels and disease labels are expanded and the corresponding repeated symptom input is as follows. We use different colors to mark the symptom transfer. The dashed box contains the original explicit symptoms, implicit symptoms, and the related disease label. $d^* =$ Allergy Rash. $d^1(d^2, d^3)$ is equal to $d^*$ if \textit{Sneezing} $($\textit{Allergy, Rash} $)$ is an available symptom, otherwise it is $\#$.}

\label{fig_reppeated}
\end{figure}
\subsection{$s$-label Augmentation} 
As mentioned above, the order of symptoms can potentially affect diagnosis accuracy. In this section, we investigate whether we can use $s$-label augmentation to provide better training signals to reduce the effects of symptom order. $s$-label augmentation is based on the assumption that determining a disease is independent of the order of symptoms. Taking advantage of this disorder, we perform additional data augmentation with symptom labels.
A sub-sequence of symptoms $s^{1:K}_I$ not only has the symptom label of the next symptom $s^{K+1}_I$, but also of the subsequent symptoms $s^{K+1:M}_I$. Instead of using the original single symptom label of $\{s^{1:N}_E, s^{1:M}_I\}$, we replace it with all possible implicit symptom labels and rewrite it as:
$$ \overbrace{s^1_I,\cdots,s^M_I}^M, \cdots, \overbrace{s^{K+1}_I,\cdots,s^M_I}^{M-K}, \cdots, \overbrace{s^{M}_I}^1, \# $$
where $\#$ is the token ignored during training. We denote the expanded symptom labels as $\widetilde{s^{1:M'}_{I}}$, where $M'$ is calculated as $M' = (1+M)\cdot M / 2 + 1$. Then the joint probabilities of $\widetilde{s^{1:M’}_{I}}$ over the symptoms are as follows:
\begin{equation}
\begin{aligned}
\begin{small}
    \mathcal{P}_{\theta}(\widetilde{s^{1:M'}_{I}}|s^{1:N}_E) = 
    \prod_{K = 1}^{M} \prod_{T = K}^{M}
    \mathcal{P}_{\theta}(s^T_I|s^{1:K-1}_I, s^{1:N}_E).
    \label{e:3}
\end{small}
\end{aligned}
\end{equation}
Similarly, the aligned disease labels can also be expanded to align with $\widetilde{s^{1:M'}_{I}}$ as the following:
$$ \overbrace{d^1_1,\cdots,d^1_M}^{M}, \cdots, \overbrace{d^{K+1}_{K+1},\cdots,d^{K+1}_{M}}^{M-K}, \cdots, \overbrace{d^{M-1}}^{1}, d^*, $$
where $d^K_T$ is given following the $d$-label assignment rules, i.e., it is $d^*$ if $s^T_I$ is the available symptom of sequence $\{s^{1:N}_E, s^{1:K-1}_I, s^T_I\}$, else it is $\#$. 
We denoted the expanded aligned disease labels as $\widetilde{d^{1:M'}}$, then the joint probabilities of $\widetilde{d^{1:M'}}$ over the symptoms are given as:
\begin{equation}
\begin{aligned}
    \mathcal{P}_{\theta}(d^*, \widetilde{d^{1:M'}} & |s^{1:M}_I, s^{1:N}_E) = \mathcal{P}_{\theta}(d^*|s^{1:M}_I, s^{1:N}_E) \\ & \prod_{K = 1}^{M} \prod_{T = K}^{M}
    \mathcal{P}_{\theta}(d^K_T|s^{1:K-1}_I, s^{1:N}_E).
    \label{e:4}
\end{aligned}
\end{equation}

\paragraph{\textbf{Repeated Symptom Input.}} Now that we have obtained the expanded symptom and disease labels, we will explain how to expand the input symptoms to align with the labels in an auto-regression generation model. The solution is straightforward, as shown in Figure \ref{fig_reppeated}, we simply repeat the symptoms as many times as their corresponding expanded symptom or disease labels. For example, in Figure \ref{fig_reppeated}, the last explicit symptom $Sneezing$ has 3 extended symptom labels $Allergy, Rash$ and $Dyspnea$, thus it will be repeated 3 times. Formally, an implicit symptom $s^K_I$ will be repeated $M-K+1$ times.
To this end, the repeated symptom inputs are represented as:
$$ \overbrace{s^N_E,\cdot\cdot,s^N_E}^{M}, \overbrace{s^1_I,\cdot\cdot,s^1_I}^{M-1}, \cdot\cdot\cdot, \overbrace{s^K_I,\cdot\cdot,s^K_I}^{M-K}, 
\cdot\cdot\cdot, \overbrace{s^{M-1}_I}^{1}, s^M_I.$$
We denote the repeated symptom input as $s^{1:M'}{rept}$, the mappings between $s^{1:M'}{rept}$ and $\widetilde{d^{1:M'}}$ as well as $s^{1:M'}_{rept}$ and $\widetilde{s^{1:M'}_I}$. In addition to the repeated symptom input, extra symptom attention is integrated to mask the redundant input symptoms. Specifically, in each multi-head attention of the Transformer block, each repeated symptom can only see itself, the explicit symptoms, and only one of the previous repeated symptoms. Formally, the representations of symptom tokens are updated in multi-head attention as: 
\begin{equation}
\begin{footnotesize}
\begin{aligned}
(s^{K'}_{I(rept)})^l
\begin{cases}
\leftarrow \text{MH-Attn}(Q=(s^{K'}_{I(rept)})^{(l-1)}, \\ KV= [(s^{X}_{I(rept)})^{(l-1)}, (s^{1:N-1}_E)^{(l-1)}]) & \\ \ \ \ \ \ \ \ \ \ \ \ \ \text{if} \ K>M \\
\leftarrow \text{MH-Attn}(Q=(s^{K'}_{I(rept)})^{(l-1)}, \\ KV=  [s^{K'}_{I(rept)})^{(l-1)}, (s^{1:N-1}_E)^{(l-1)}])    & \text{else,}
\end{cases}
\end{aligned}
\end{footnotesize}
\end{equation}
where $Q, K, V$ are the query, key, and value in multi-head attention respectively. $[.]$ represents the concatenation along the symptom sequence dimension, $(s^{K'}_I(rept))^l$ indicates the $l^{th}$ Transformer block layer output of the ${K'}^{th}$ repeated symptoms input, and $X = \{\frac{(K+1)(2\cdot M-K)}{2}\}, K = 0, 1, \cdots, \text{such that} \ X < K'$. 
Significantly, $s^{M'}_{I(rept)}$ is anchored for the final diagnosis with only disease label $d^*$, its representation of the $l^{th}$ layer is updated in multi-head attention as $\text{MH-Attn}(Q=(s^{M'}_{I(rept)})^{(l-1)}, KV= [(s^{\{\frac{(K+1)(2\cdot M-K)}{2}, M'\}}_{I(rept)})^{(l-1)}, (s^{1:N-1}_E)^{(l-1)}])$, where $K = 0, \cdot\cdot\cdot, M$.


The symptom attention mechanism is a key aspect of the repeated symptom input, functioning as a form of data augmentation akin to dropout noise in accomplished by Transformer training \cite{vaswani2017attention}. This is achieved by placing dropout masks on both the fully-connected feed-forward layers and the attention probabilities. During repeated symptom generation, dropout noise serves as data augmentation by independently sampling dropout masks \cite{gao2021simcse}. This process involves feeding the same samples to the decoder multiple times and performing data augmentation.

It is important to note that the key distinction between our approach and a simple symptom permutation method lies in the dropout noise applied to the explicit symptoms. In the permutation approach, different representations of the explicit symptoms are used as context due to varying dropouts applied to each separate augmented sample. In contrast, our method with the addition of the $S$-label utilizes the same representations of the explicit symptoms as context for all subsequent implicit symptoms. Maintaining the stability of the explicit symptoms’ representation is essential for the model to learn more effectively.


\paragraph{\textbf{Collaborative Generation Loss.}} We follow the auto-regression framework in \cite{vaswani2017attention} and take a cross-entropy objective to learn the expanded labels $\widetilde{d^{1:M'}}$ and $\widetilde{s^{1:M'}_{I}}$ jointly as minimizing the following loss:
\begin{equation}
\setlength{\abovedisplayskip}{3pt}
\setlength{\belowdisplayskip}{3pt}
\begin{aligned}
    \mathcal{L}_{\theta} & = - \sum_{K=1}^{M} \sum_{T=K}^M  W^K \cdot \left[ \log \mathcal{P}_{\theta} (s^T_I|s^{1:{K-1}}, s^N_E) + \right. \\& \left. \log \mathcal{P}_{\theta} (d^K_T|s^{1:{K-1}}, s^N_E) \right] + \log \mathcal{P}_{\theta} (d^*|s^{1:M}, s^N_E) \\
    & =- \sum_{K'=1}^{M'} W^{K'} \cdot \left[ \log \mathcal{P}_{\theta} (\widetilde{s^{K'}_{I}}|s^{1:K'-1}_{rept}, s^N_E) \right. \\& \ \ \ \ \ \ \ \left. + \log \mathcal{P}_{\theta} (\widetilde{d^{K'}}|s^{1:K'-1}_{rept}, s^N_E) \right],
    \label{e:5}
\end{aligned}
\end{equation}
where $W^K = \frac{1}{M-T+1}$ is the weight added to normalize the $\sum_{T=K}^M \log \mathcal{P}_{\theta} (s^T_I|s^{1:{K-1}}, s^N_E)$, and $W^{K'}$ is the weight to normalize the losses of the expanded labels in the repeated symptom input. It can be set as $\frac{1}{M-n-1}, \text{where} \ \frac{n(2 \cdot M - n + 1)}{2} < K' \leq \frac{(n+1)(2 \cdot M)}{2}$.

\section{Experiments}

\begin{table*}[!htbp]
\renewcommand{\arraystretch}{.75}
\small
\centering
\setlength{\tabcolsep}{0.5mm}{
\begin{tabular}{l @{\hskip 1.7mm} cccc @{\hskip 1.5mm} cccc @{\hskip 1.5mm} cccc @{\hskip 1.5mm} cccc}
\toprule 
\multirow{2}{*}{Model}
&\multicolumn{4}{c}{Dxy}
&\multicolumn{4}{c}{Muzhi}
&\multicolumn{4}{c}{Muzhi-2}
&\multicolumn{4}{c}{Ped}\\ 
\cmidrule(l){2-5} \cmidrule(l){6-9} \cmidrule(l){10-13} \cmidrule(l){14-17}

& $Ac \uparrow$ &$Rc \uparrow$  &$Cs \uparrow$ &$T \downarrow$
& $Ac \uparrow$ &$Rc \uparrow$  &$Cs \uparrow$ &$T \downarrow$
& $Ac \uparrow$ &$Rc \uparrow$  &$Cs \uparrow$ &$T \downarrow$
& $Ac \uparrow$ &$Rc \uparrow$  &$Cs \uparrow$ &$T \downarrow$
\\
\midrule
BERT$_{full}$   
        &0.83 & $-$ & $-$ & $-$
        &0.73 & $-$ & $-$ & $-$
        &0.65 & $-$ & $-$ & $-$
        &0.34 & $-$ & $-$ & $-$ \\
        \midrule
DQN$\dagger$ 
        & 0.72 & 0.32& 0.44 &{\color{blue}{\bf{2.4}}}
        & 0.69 & 0.30& 0.42&{\color{blue}{\bf{3.1}}}
        & $-$ & $-$ & $-$ & $-$
        & $-$ & $-$ & $-$ & $-$\\
        
PPO  
        & 0.78 & 0.31& 0.45&4.7
        & 0.72& 0.23& 0.35 & 4.5
        & 0.59& 0.18& 0.28&{\color{blue}{\bf{6.6}}}
        & 0.36& 0.28 & 0.31&{\color{blue}{\bf{11.9}}}\\
        
H-DQN$\dagger$ 
        & 0.70 & 0.16& 0.26 &{\color{blue}{\bf{2.4}}}
        & 0.69&0.28 & 0.40&3.5
        & $-$ & $-$ & $-$ & $-$
        & $-$ & $-$ & $-$ & $-$\\
        
Diaformer$\dagger,\ddagger$ 
        & 0.83& 0.83& 0.83 &13.1
        & 0.74& 0.75& 0.75 &15.3
        & 0.64& 0.61& 0.63 &11.5
        &0.51 &0.83 &0.63 &14.3\\
\midrule        
GPT-2$_{dis}$    
        &0.83 &0.90 &0.87 &15.8
        &0.72 &0.74 &0.73 &17.6
        &0.63 &0.62 &0.62 &16.2
        &0.48 &0.81 &0.60 &18.8\\
CoAD$_{w/o \ d}$       
        &0.83 &{\color{blue}{\bf{0.93}}} &0.88& 15.1
        &0.72 &{\color{blue}{\bf{0.83}}} &0.77& 16.2
        &{\color{blue}{\bf{0.65}}}& 0.67 &{\color{blue}{\bf{0.66}}}& 16.9
        & 0.52& 0.87& 0.54& 17.1\\
CoAD$_{w/o \ s}$       
        &0.84 &0.84 &0.84 &14.9
        &0.70 &0.81 & 0.75& 15.2
        &0.64 &0.67 &0.65 &16.2
        &0.52 &0.84 &0.64 &18.3\\
CoAD$^*$ 
        &{\color{blue}{\bf{0.85}}}&{\color{blue}{\bf{0.93}}}&{\color{blue}{\bf{0.89}}} & 10.5
        &{\color{blue}{\bf{0.75}}}&{\color{blue}{\bf{0.83}}}&{\color{blue}{\bf{0.79}}} & 13.4 
        & {\color{blue}{\bf{0.65}}}& {\color{blue}{\bf{0.68}}}& {\color{blue}{\bf{0.66}}}& 13.8
        &{\color{blue}{\bf{0.53}}} &{\color{blue}{\bf{0.92}}} &{\color{blue}{\bf{0.67}}} &15.4\\

\bottomrule 
\end{tabular}
}
\caption{\label{tab: main} The average disease accuracy ($Ac$), implicit symptom recall ($Rc$), combined score ($Cs$), and number of turns ($T$) on the four datasets with 20 limited turns. $\dagger$: The results on Dxy and Muzhi are reported by \citet{chen2022diaformer}. $\ddagger$: The results on Muzhi-2 and Ped are obtained by running the authors' released code. The symbol $^*$ signifies a significant level of $p < 0.05$ when compared to both the baselines and CoAD’s variants.} 
\end{table*}

\paragraph{\textbf{Datasets.}}
We evaluate our model on three public datasets: Dxy \cite{xu2019end}, MuZhi \cite{wei2018task}, MuZhi-2, and one in private---Ped. We only consider datasets collected from real clinical practice and exclude the Synthetic dataset \cite{liao2020task}. Additional detail of the data is given in Appendix \ref{e:1}. 

\paragraph{\textbf{Baselines.}}
In order to evaluate the effectiveness of RL-based AD models, we utilize several established techniques as benchmark models including: \textbf{DQN} \cite{wei2018task} agent which uses Deep Q-Network to learn an agent that chooses a symptom to inquire about or outputs the disease diagnosis; \textbf{PPO} \cite{schulman2017proximal} agent which is a Proximal Policy Optimization \cite{schulman2017proximal} based agent that learns diagnosis model with a GPT-2 backbone, and Hierarchical DQN \textbf{(HDQN)} \cite{liao2020task} agent which is a DQN-based hierarchical policy that has a master selecting the disease and a separate worker for each disease to inquire symptoms. Additionally, we include the current leading Transformer-based model, \textbf{Diaformer} \cite{chen2022diaformer} agent which formulates AD as a sequence generation task and learns a Transformer-based model to learn the generation of symptom inquiry and disease diagnosis. 

\paragraph{\textbf{Ablation Study.}}
Some CoAD's variants includes \textbf{GPT-2$_{dis}$} agent which is based on GPT-2 \cite{radford2019language} that generates symptoms in an auto-regressive manner and predicts the disease at the end; \textbf{CoAD$_{w/o \ s}$ } agent which is a variant of CoAD that is trained without $s$-label using the same hyperparameters of CoAD; \textbf{CoAD$_{w/o \ d}$ } agent which is a variant of CoAD that is trained without $d$-label using the same hyperparameters of CoAD. Meanwhile, a \textbf{BERT$_{full}$} \cite{devlin-etal-2019-bert} agent which uses BERT as the backbone to classify disease given the full ground truth symptoms is compared. 

\subsection{Experiment Setup}
\paragraph{\textbf{Evaluation Metrics.}}
We evaluate the performance of the models using four metrics: implicit symptom recall, disease accuracy, average inquiry turns, and a combined score. The combined score is calculated as the harmonic mean of disease accuracy and implicit symptom recall, providing an overall measure of the model's effectiveness in both disease diagnosis and symptom inquiry. To ensure the reliability of the findings, all evaluation results are derived from the average of five distinct groups.

\paragraph{\textbf{Training Setting.}}
To gain insight into the behavior of the models under different conditions, we conduct experiments in both limited turn and fixed turn modes. In the fixed turn mode, the models are required to inquire about symptoms within a fixed number of turns. In the limited turn mode, there is a maximum turn limit for symptom inquiry, but the models are allowed to stop before reaching the limit. To ensure a fair comparison with the previous state-of-the-art Diaformer \cite{chen2022diaformer}, we use the same turn numbers (5, 10, 15, and 20) in our experiments. Additional implementation is shown in Appendix \ref{appdx:implementation}.  
\begin{table*}[!htbp]
\renewcommand{\arraystretch}{0.75}
\small
\centering
\centering
\setlength{\tabcolsep}{0.5 mm}
{
\begin{tabular}{c @{\hskip 1mm} l cccc @{\hskip 1mm} cccc @{\hskip 1mm}cccc@{\hskip 1mm} cccc}
\toprule 
\multirow{2}{*}{\makecell[c]{Max\\Turns}}&\multirow{2}{*}{Model}
&\multicolumn{4}{c}{Dxy}
&\multicolumn{4}{c}{Muzhi}
&\multicolumn{4}{c}{Muzhi-2}
&\multicolumn{4}{c}{Ped}\\ 
\cmidrule(l){3-6} \cmidrule(l){7-10} \cmidrule(l){11-14} \cmidrule(l){15-18}

& & $Ac \uparrow$ &$Rc \uparrow$  &$Cs \uparrow$ &$T \downarrow$
& $Ac \uparrow$ &$Rc \uparrow$  &$Cs \uparrow$ &$T \downarrow$
& $Ac \uparrow$ &$Rc \uparrow$  &$Cs \uparrow$ &$T \downarrow$
& $Ac \uparrow$ &$Rc \uparrow$  &$Cs \uparrow$ &$T \downarrow$
\\
\midrule
\multirow{8}{*}{5} 
&DQN$\dagger$ 
        &0.65 &0.31 &0.42 &2.5
        &0.64 &0.29 &0.40 &2.9 
        &$-$   &$-$   &$-$   &$-$
        &$-$   &$-$   &$-$   &$-$\\
        
&PPO  
        &0.73 &0.24 &0.36 &2.6
        &0.72 &0.18 &0.29 &{\color{blue}{\bf{2.1}}}
        &0.58 &0.14 &0.22 &{\color{blue}{\bf{3.2}}}
        &0.39 &0.16 &0.22 &4.2\\
        
&H-DQN$\dagger$ 
        &0.70 &0.15 &0.25 &{\color{blue}{\bf{1.9}}}
        &0.68 &0.29 &0.40 &2.9 
        &$-$   &$-$   &$-$   &$-$
        &$-$   &$-$   &$-$   &$-$\\
        
&Diaformer$\dagger, \ddagger$ 
        &0.77 &0.55 &0.64 &4.8
        &{\color{blue}{\bf{0.72}}} &0.47 &0.57 &5.0 
        &{\color{blue}{\bf{0.60}}} &{\color{blue}{\bf{0.39}}} &{\color{blue}{\bf{0.47}}} &4.9
        &0.46 &0.58 &0.49 &4.4\\
\cmidrule(l){2-18}      
&GPT-2$_{dis}$
        &0.72 &0.58 &0.64 &4.8
        &0.70 &0.51 &0.60 &5.0
        &0.57 &0.33 &0.42 &5.0
        &0.40 &0.42 &0.41 &{\color{blue}{\bf{3.3}}}\\
&CoAD$_{w/o \ d}$       
        &0.75 &{\color{blue}{\bf{0.59}}} &{\color{blue}{\bf{0.65}}} &5.0
        &0.70 &0.52 &0.60 &5.0
        &0.58 &0.38 &0.42 &5.0
        &0.42 &0.60 &0.49 &4.4\\
&CoAD$_{w/o \ s}$       
        &0.75&0.56&0.64&4.4
        &0.70&0.51&0.59&5.0
        &0.58&0.34&0.43&5.0
        &0.44&0.58&0.50&4.5\\
&CoAD$^*$ 
        &{\color{blue}{\bf{0.79}}} &0.56 &{\color{blue}{\bf{0.65}}} &4.9
        &{\color{blue}{\bf{0.72}}} &{\color{blue}{\bf{0.53}}} &{\color{blue}{\bf{0.61}}} &4.8
        &{\color{blue}{\bf{0.60}}} &{\color{blue}{\bf{0.39}}} &{\color{blue}{\bf{0.47}}} &5.0
        &{\color{blue}{\bf{0.47}}} &{\color{blue}{\bf{0.61}}} &{\color{blue}{\bf{0.53}}} &4.6\\

\bottomrule 
\multirow{8}{*}{10} 
&DQN$\dagger$ 
        & 0.72 & 0.32& 0.44 & 2.7
        & 0.68 & 0.30& 0.41& 3.0
        &$-$   &$-$   &$-$   &$-$
        &$-$   &$-$   &$-$   &$-$\\
        
&PPO  
        &0.75 &0.30 &0.43 &3.6
        &0.68 &0.20 &0.31 &{\color{blue}{\bf{2.4}}}
        &0.61 &0.15 &0.24 &{\color{blue}{\bf{3.5}}}
        &0.37 &0.28 &0.32 &7.8\\
        
&H-DQN$\dagger$ 
        &0.72 &0.16 &0.26 &{\color{blue}{\bf{2.3}}}
        &0.70 &0.27 &0.39 &3.3
        &$-$   &$-$   &$-$   &$-$
        &$-$   &$-$   &$-$   &$-$\\
        
&Diaformer$\dagger, \ddagger$ 
        &0.81 &0.78 &0.79 &9.6
        &{\color{blue}{\bf{0.73}}} &0.66 &0.69 &9.8
        &{\color{blue}{\bf{0.62}}} &{\color{blue}{\bf{0.58}}} &0.59 &9.8
        &0.50 &0.76 &0.58 &7.8\\
\cmidrule(l){2-18}         
&GPT-2$_{dis}$  
        &0.80 &0.79 &0.79 &9.7
        &0.71 &0.70 &0.70 &9.7
        &0.60 &0.50 &0.54 &10.0
        &0.42 &0.45 &0.43 &{\color{blue}{\bf{4.2}}}\\
&CoAD$_{w/o \ d}$       
        &0.82 &{\color{blue}{\bf{0.83}}} &0.82 &9.7
        &0.71 &{\color{blue}{\bf{0.71}}} &{\color{blue}{\bf{0.71}}} &9.6
        &{\color{blue}{\bf{0.62}}} &0.51 &0.56 &10.0
        &0.44 &0.75 &0.55 &5.0\\
&CoAD$_{w/o \ s}$       
        &0.78&0.79&0.78&9.6
        &{\color{blue}{\bf{0.73}}}&0.70 &{\color{blue}{\bf{0.71}}} &9.5
        &0.60 &0.51 &0.56&9.8
        &0.46&0.71&0.56&7.4\\
&CoAD$^*$ 
        &{\color{blue}{\bf{0.85}}} &0.80 &{\color{blue}{\bf{0.83}}} &9.3
        &{\color{blue}{\bf{0.73}}} &0.70 & {\color{blue}{\bf{0.71}}} &9.4
        &{\color{blue}{\bf{0.62}}} &{\color{blue}{\bf{0.58}}} &{\color{blue}{\bf{0.60}}} &9.9
        &{\color{blue}{\bf{0.51}}} &{\color{blue}{\bf{0.78}}} &{\color{blue}{\bf{0.62}}} &8.1\\

\bottomrule 
\multirow{8}{*}{15} 
&DQN$\dagger$ 
        & 0.71 & 0.32& 0.44 & 2.7
        & 0.68 & 0.30& 0.41& {\color{blue}{\bf{3.0}}}  
        &$-$   &$-$   &$-$   &$-$
        &$-$   &$-$   &$-$   &$-$\\
        
&PPO  
        &0.77 &0.30 &0.43 &4.1
        &0.74 &0.24 &0.37 &4.0
        &0.62 &0.12 &0.21 &{\color{blue}{\bf{3.5}}}
        &0.37 &0.23 &0.28 &{\color{blue}{\bf{6.5}}}\\
        
&H-DQN$\dagger$ 
        &0.72 &0.16 &0.26 &{\color{blue}{\bf{2.3}}}
        &0.70 &0.27 &0.39 &3.4
        &$-$   &$-$   &$-$   &$-$
        &$-$   &$-$   &$-$   &$-$\\
        
&Diaformer$\dagger, \ddagger$ 
        &0.83 &0.83 &0.83 &12.4
        &0.74 &0.73 &0.69 &13.8
        &0.62 &{\color{blue}{\bf{0.64}}} &0.62 &12.4
        &0.50 &0.81 &0.60 &9.3\\
\cmidrule(l){2-18}         
&GPT-2$_{dis}$
        &0.83 &0.85 &0.84 &13.5
        &0.68 &0.76 &0.72 &14.8
        &0.60 &0.58 &0.59 &15.0
        &0.41&0.52 &0.46 &9.2\\
&CoAD$_{w/o \ d}$       
        &0.84 &{\color{blue}{\bf{0.91}}} &{\color{blue}{\bf{0.88}}} &14.5
        &0.72 &{\color{blue}{\bf{0.80}}} &0.76 &14.3
        &0.63 &0.63 &{\color{blue}{\bf{0.64}}} &15.0
        &0.45 &0.84 &0.59 &9.6\\
&CoAD$_{w/o \ s}$       
        &0.82&0.89&0.85&14.3
        &0.71&{\color{blue}{\bf{0.80}}} &0.75&14.6
        &0.63&0.62&0.63&14.7
        &0.46&0.82&0.59&8.6\\
&CoAD$^*$ 
        &{\color{blue}{\bf{0.85}}} &0.90 &{\color{blue}{\bf{0.88}}} &13.5
        &{\color{blue}{\bf{0.75}}} &{\color{blue}{\bf{0.80}}} &{\color{blue}{\bf{0.77}}} &13.6
        &{\color{blue}{\bf{0.64}}} &0.62 &{\color{blue}{\bf{0.64}}} &15.0
        &{\color{blue}{\bf{0.51}}} &{\color{blue}{\bf{0.86}}} &{\color{blue}{\bf{0.64}}} &10.2\\

\bottomrule 
\end{tabular}
\caption{\label{tab: ablative}
The results on the four datasets in three limited turn settings are presented. The notations $Ac$, $Rc$, $Cs$, $T$ are defined in Table \ref{tab: main}. The results on Dxy and Muzhi were reported by \citet{chen2022diaformer} and marked with $\dagger$. The results on Muzhi-2 and Ped were obtained by running the authors' released code and are marked with $\ddagger$. The symbol $^*$ signifies a significant level of $p < 0.05$ when compared to both the baselines and CoAD’s variants.
} 
}
\end{table*}

\paragraph{\textbf{Main results.}}
Table \ref{tab: main} shows the evaluation results on four AD number of 20 turns. For the models DQN and H-DQN, we cite the results from \cite{chen2022diaformer} where only the results for Dxy and Muzhi are available. The main results indicate that the proposed CoAD model achieves the highest disease accuracy, symptom recall, and combined score on all datasets, demonstrating the effectiveness of the $d$-label and $s$-label for AD. Compared to the previous state-of-the-art model, Diaformer, CoAD shows a significant improvement in disease accuracy, for example, in the Ped dataset the gain of CoAD is 3.92\%. The improvement in disease accuracy is a result of the combined effects of the $d$-label and $s$-label. Specifically, the $s$-label helps CoAD capture symptom relationships in different sequences, while the $d$-label helps CoAD generalize better to unseen symptom sequences. Furthermore, CoAD achieves a substantial improvement in symptom recall in all datasets over Diaformer with at least 10\% improvement, highlighting the potential of repeated symptom input in improving the model's ability to inquire about appropriate symptoms during diagnosis. Notably, both Muzhi-2 and Ped contain negative symptom statuses, adding complexity to the symptom sequence input and providing more challenges to learning the relationships between symptoms and the target disease. However, both $d$-label alignment and $s$-label augmentation consistently bring improvements regardless of the setting.

It is worth noting that all the reinforcement learning (RL) methods present poor performance in terms of symptom recall, which is not unexpected as they tend to inquire about a limited number of symptoms and stop at early turns. The early stop of RL-based methods can be attributed to intermediate negative rewards, which incentivize the model to end in as few turns as possible for efficiency. On the other hand, generation-based methods are generally capable of probing more symptoms compared to RL-based methods. Notably, CoAD achieves the best performance with shorter turns in comparison to other generation-based methods, demonstrating the effectiveness and efficiency of CoAD's diagnostic logic.

Finally, the BERT$_{full}$ model is trained to predict the disease based on the sequence of ground truth symptoms. Intuitively, we expected this model to provide the theoretical upper bound for disease accuracy. However, both Diaformer and CoAD outperform BERT$_{full}$ in all datasets except Muzhi-2. This is not surprising as there can be irrelevant symptoms that negatively impact the accuracy of decision-making for disease diagnosis. Additionally, BERT$_{full}$ model lacks the ability to distinguish the symptoms that are relevant to the final diagnosed disease. 


\paragraph{\textbf{Diagnosis with smaller limited turns.}}
Table \ref{tab: ablative} presents the results of automatic diagnosis with smaller limited turns (i.e., 5, 10, 15) on the four datasets. The results indicate that CoAD consistently outperforms Diaformer in terms of disease accuracy, symptom recall, and combined score, demonstrating the robustness of the proposed methods under different limited turn settings. Specifically, CoAD shows significant improvements in disease accuracy and symptom recall compared to previous results. As the allowed turns increase, we observe a monotonic improvement in both disease accuracy and symptom recall, indicating that CoAD can provide a better-quality diagnosis with more turns.

\paragraph{\textbf{Diagnosis with fixed turns.}}

For fairer comparisons, we evaluate the models under a fixed number of turns and the results are presented in Figure \ref{fig:fix_turn} in Appendix \ref{appdx:fix_turn}. We choose the fixed turns as 5, 10, and 15. The DQN-based method \cite{wei2018task} is not considered as it uses a single action space for both disease and symptoms, making it only suitable for limited turn settings. Overall, the results indicate that CoAD achieves the best performance across all evaluation metrics on all datasets with different fixed turns (except for the disease accuracy of Muzhi with a fixed turn of 15). Notably, the improvement is more substantial in shorter turns (8.2\% in 15 turns vs 2.5\% in 5 turns increase on average over Diaformer), showcasing the strength of CoAD in real-life deployment scenarios where efficiency is crucial.

In terms of symptom recall, CoAD consistently outperforms the other models by a wide margin. The largest improvement is 24\% on the Muzhi dataset with five fixed turns over Diaformer. In contrast, the improvement in disease accuracy is relatively modest. The significant improvement in symptom recall aligns with expectations, as the $d$-label augmentation encourages CoAD to explore the relationship between intermediate symptoms and the final disease during training, allowing CoAD to inquire about the most relevant symptoms for distinguishing different diseases. As a result, CoAD is able to make correct diagnostic decisions even with insufficient symptoms. Therefore, when CoAD is forced to inquire about more symptoms in the fixed turns, the additional symptoms provide less value in diagnosis compared to other models.


\paragraph{\textbf{Ablation Studies.}}

To further understand the contributions of the different components of CoAD, we conduct a series of ablation studies to isolate the effects of $s$-label augmentation and $d$-label augmentation. In Table \ref{tab: main} and Table \ref{tab: ablative}, we can observe that CoAD${w/o \ d}$ consistently improves symptom recall across all datasets and different limited turns. These results indicate that the $s$-label is effective in guiding the model to inquire about informative symptoms, leading to more accurate diagnoses and, as a result, better disease accuracy in most settings. On the other hand, CoAD${w/o \ s}$ improves disease accuracy over GPT-2$_{dis}$ in most settings (e.g., an improvement of 12\% in the Ped dataset), highlighting the effectiveness of $d$-label augmentation. Furthermore, the combination of $d$-label and $s$-label is beneficial as CoAD achieves the best disease accuracy and combined score in all settings, and better symptom recall in most cases.

\section{Conclusions}
This paper introduces CoAD, a symptom and disease co-generation framework, which significantly improves the state-of-the-art in symptom prediction and disease determination for automatic diagnosis. CoAD addresses the discrepancy between visible symptoms during training and generation through disease label alignment, mitigates the impact of symptom order on diagnosis through symptom label augmentation, and utilizes a repeated symptoms input schema to enable the model to simultaneously learn aligned diseases and expanded symptoms efficiently. CoAD presents a novel approach to data augmentation by reusing labels and text input, and it can be extended to other joint learning tasks for generation and classification.

\section*{Limitations}
In this work, we have identified two key limitations of CoAD that can be further examined in future research. The first limitation is that CoAD only allows for the querying of one symptom at a time, making it unsuitable for scenarios where multiple symptoms are present. However, CoAD has superior performance in the main metrics for automatic diagnosis. To relieve this limitation, potential solutions include relaxing symptom feedback conditions and allowing the model to produce symptoms sequentially until a stop signal is encountered or querying the top K symptoms in a single turn. 

Additionally, CoAD has some restrictions on input format, requiring standardized symptoms and values. To make it more applicable to end-to-end settings, an natural language understanding module (NLU) is required to parse plain text and obtain the input symptom sequence, and a natural language generation (NLG) module is needed to translate the predicted symptom or disease to text. The ultimate goal of automatic diagnosis is to support the dialogue between doctors and patients, after CoAD determines the symptom or disease, rule-based NLU and NLG modules can help to achieve the text to text communication.

\section*{Ethics Statement}

Our work adheres to the ACL Ethics Policy. This paper aims to investigate generative model-based approaches for learning automatic diagnostic logic, with the objective of reducing the burden on doctors and promoting the advancement of automatic diagnosis systems. It is crucial to emphasize that the proposed methods are designed solely for research purposes and are not suitable for direct clinical application due to the potential risks associated with the misuse of automatic diagnosis systems.

It is important to note that the introduced dataset (Ped) was sourced from genuine electronic medical records, with all patient privacy-related information meticulously eliminated. To ensure data privacy and security, we performed a comprehensive manual review of the dataset, confirming that it contains no identifiable or offensive pieces of information within the experimental dataset.








\section*{Acknowledgements}
We appreciate the constructive and insightful comments provided by the anonymous reviewers. This research work is partially supported by CUHK under Project No. 3230377.

\bibliography{anthology,custom,www}

\begin{thebibliography}{15}
\expandafter\ifx\csname natexlab\endcsname\relax\def\natexlab#1{#1}\fi

\bibitem[{Chen et~al.(2022)Chen, Li, Chen, Zhou, and Liu}]{chen2022diaformer}
Junying Chen, Dongfang Li, Qingcai Chen, Wenxiu Zhou, and Xin Liu. 2022.
\newblock Diaformer: Automatic diagnosis via symptoms sequence generation.
\newblock In \emph{Proceedings of the AAAI Conference on Artificial
  Intelligence}, volume~36, pages 4432--4440.

\bibitem[{Devlin et~al.(2019)Devlin, Chang, Lee, and
  Toutanova}]{devlin-etal-2019-bert}
Jacob Devlin, Ming-Wei Chang, Kenton Lee, and Kristina Toutanova. 2019.
\newblock \href {https://doi.org/10.18653/v1/N19-1423} {{BERT}: Pre-training of
  deep bidirectional transformers for language understanding}.
\newblock In \emph{Proceedings of the 2019 Conference of the North {A}merican
  Chapter of the Association for Computational Linguistics: Human Language
  Technologies, Volume 1 (Long and Short Papers)}, pages 4171--4186,
  Minneapolis, Minnesota. Association for Computational Linguistics.

\bibitem[{Gao et~al.(2021)Gao, Yao, and Chen}]{gao2021simcse}
Tianyu Gao, Xingcheng Yao, and Danqi Chen. 2021.
\newblock Simcse: Simple contrastive learning of sentence embeddings.
\newblock In \emph{Proceedings of the 2021 Conference on Empirical Methods in
  Natural Language Processing}, pages 6894--6910.

\bibitem[{Kao et~al.(2018)Kao, Tang, and Chang}]{kao2018context}
Hao-Cheng Kao, Kai-Fu Tang, and Edward Chang. 2018.
\newblock Context-aware symptom checking for disease diagnosis using
  hierarchical reinforcement learning.
\newblock In \emph{Proceedings of the AAAI Conference on Artificial
  Intelligence}, volume~32.

\bibitem[{Liao et~al.(2020)Liao, Liu, Wei, Peng, Chen, Sun, and
  Huang}]{liao2020task}
Kangenbei Liao, Qianlong Liu, Zhongyu Wei, Baolin Peng, Qin Chen, Weijian Sun,
  and Xuanjing Huang. 2020.
\newblock Task-oriented dialogue system for automatic disease diagnosis via
  hierarchical reinforcement learning.
\newblock \emph{arXiv preprint arXiv:2004.14254}.

\bibitem[{Lin et~al.(2019)Lin, He, Chen, Tou, Wei, and Chen}]{lin2019enhancing}
Xinzhu Lin, Xiahui He, Qin Chen, Huaixiao Tou, Zhongyu Wei, and Ting Chen.
  2019.
\newblock Enhancing dialogue symptom diagnosis with global attention and
  symptom graph.
\newblock In \emph{Proceedings of the 2019 Conference on Empirical Methods in
  Natural Language Processing and the 9th International Joint Conference on
  Natural Language Processing}, pages 5033--5042.

\bibitem[{Peng et~al.(2018)Peng, Tang, Lin, and Chang}]{peng2018refuel}
Yu-Shao Peng, Kai-Fu Tang, Hsuan-Tien Lin, and Edward Chang. 2018.
\newblock Refuel: Exploring sparse features in deep reinforcement learning for
  fast disease diagnosis.
\newblock In \emph{Advances in Neural Information Processing Systems},
  volume~31.

\bibitem[{Radford et~al.(2019)Radford, Wu, Child, Luan, Amodei, and
  Sutskever}]{radford2019language}
Alec Radford, Jeffrey Wu, Rewon Child, David Luan, Dario Amodei, and Ilya
  Sutskever. 2019.
\newblock Language models are unsupervised multitask learners.
\newblock \emph{OpenAI blog}, 1(8):9.

\bibitem[{Schulman et~al.(2017)Schulman, Wolski, Dhariwal, Radford, and
  Klimov}]{schulman2017proximal}
John Schulman, Filip Wolski, Prafulla Dhariwal, Alec Radford, and Oleg Klimov.
  2017.
\newblock Proximal {{Policy Optimization Algorithms}}.
\newblock \emph{arXiv:1707.06347 [cs]}.

\bibitem[{Vaswani et~al.(2017)Vaswani, Shazeer, Parmar, Uszkoreit, Jones,
  Gomez, Kaiser, and Polosukhin}]{vaswani2017attention}
Ashish Vaswani, Noam Shazeer, Niki Parmar, Jakob Uszkoreit, Llion Jones,
  Aidan~N Gomez, {\L}ukasz Kaiser, and Illia Polosukhin. 2017.
\newblock Attention is all you need.
\newblock In \emph{Advances in Neural Information Processing Systems},
  volume~30.

\bibitem[{Wei et~al.(2018)Wei, Liu, Peng, Tou, Chen, Huang, Wong, and
  Dai}]{wei2018task}
Zhongyu Wei, Qianlong Liu, Baolin Peng, Huaixiao Tou, Ting Chen, Xuan-Jing
  Huang, Kam-Fai Wong, and Xiang Dai. 2018.
\newblock Task-oriented dialogue system for automatic diagnosis.
\newblock In \emph{Proceedings of the 56th Annual Meeting of the Association
  for Computational Linguistics (Volume 2: Short Papers)}, pages 201--207.

\bibitem[{{World Health Organization}(2016)}]{world2016health}
{World Health Organization}. 2016.
\newblock \href {https://apps.who.int/iris/handle/10665/250330} {Health
  workforce requirements for universal health coverage and the sustainable
  development goals.}

\bibitem[{Xia et~al.(2020)Xia, Zhou, Shi, Lu, and Huang}]{xia2020generative}
Yuan Xia, Jingbo Zhou, Zhenhui Shi, Chao Lu, and Haifeng Huang. 2020.
\newblock Generative adversarial regularized mutual information policy gradient
  framework for automatic diagnosis.
\newblock In \emph{Proceedings of the AAAI Conference on Artificial
  Intelligence}, volume~34, pages 1062--1069.

\bibitem[{Xu et~al.(2019)Xu, Zhou, Gong, Liang, Tang, and Lin}]{xu2019end}
Lin Xu, Qixian Zhou, Ke~Gong, Xiaodan Liang, Jianheng Tang, and Liang Lin.
  2019.
\newblock End-to-end knowledge-routed relational dialogue system for automatic
  diagnosis.
\newblock In \emph{Proceedings of the AAAI Conference on Artificial
  Intelligence}, volume~33, pages 7346--7353.

\bibitem[{Zhong et~al.(2022)Zhong, Liao, Chen, Liu, Peng, Huang, Peng, and
  Wei}]{10.1093/bioinformatics/btac408}
Cheng Zhong, Kangenbei Liao, Wei Chen, Qianlong Liu, Baolin Peng, Xuanjing
  Huang, Jiajie Peng, and Zhongyu Wei. 2022.
\newblock \href {https://doi.org/10.1093/bioinformatics/btac408} {{Hierarchical
  reinforcement learning for automatic disease diagnosis}}.
\newblock \emph{Bioinformatics}.

\end{thebibliography}
\bibliographystyle{acl_natbib}
\clearpage

\appendix
\section{Dataset}
The statistics of these datasets can be found in Table \ref{tab:dataset}. 
\label{appdx:data}
\begin{table}[htbp]
\small
\centering
\setlength{\tabcolsep}{0.5mm}{
\begin{tabular}{ccccc}
   \toprule
   Dataset &Dxy  &Muzhi  &Muzhi-2 &Ped  \\
   \midrule
   $\#$ Disease   &5     &4    &6    &44    \\
   $\#$ Symptom   &41    &66   &347  &273   \\
   Symptom type   &True    &True   &True/False  &True/False   \\
   Average length &4.7   &5.7  &9.9   &9.6  \\
   $\#$ Training  &421   &568  &1882  &5000  \\
   $\#$ Test      &104   &142  &165   &1000  \\
   \bottomrule
\end{tabular}
}
\caption{The statistics of the four datasets.}
\label{tab:dataset}
\end{table}
We evaluate our model on four datasets:
\begin{itemize}
\item \textbf{Dxy} dataset was collected from a popular Chinese healthcare website\footnote{https://dxy.com/}, where users can communicate with doctors online. The dataset records the interactions between patients and doctors, where the doctor collects more symptoms from the patient based on the initial reported symptoms. At the end of the interaction, the patient receives a diagnosis. This dataset only includes positive symptoms.
\item \textbf{MuZhi} dataset was collected from another popular Chinese online healthcare website.\footnote{http://muzhi.baidu.com} The setup of this website is similar to Dxy, and it also only includes positive symptoms.
\item \textbf{MuZhi-2} dataset was collected from the first intelligent interactive diagnosis and treatment competition (CCL 2021).\footnote{http://www.fudan-disc.com/sharedtask/imcs21/index.html} Each record contains the symptoms and exams explicitly mentioned in the patients' chief complaints and in the conversations between the patient and the doctor. The symptoms recorded in this dataset include both positive and negative symptoms.
\item \textbf{Ped} dataset was extracted from more than 6000 pediatric electronic medical records. Each piece of data consists of the symptoms mentioned in a record and the disease diagnosis given by the doctor. Similar to Muzhi-2, it includes both positive and negative symptoms.
\end{itemize}
\section{Implementation}
\label{appdx:implementation}
\begin{table}[htbp]
    \centering
    \begin{tabular}{c cc}
        \toprule
         Dataset & Learning rate & Batch size  \\
         \midrule
        Dxy & $5 \times 10^{-6} $ & 64 \\
        Muzhi & $1 \times 10^{-6} $& 64\\
        Muzhi-2 & $5 \times 10^{-6} $& 32\\
        Ped & $1 \times 10^{-6} $& 32\\
        \bottomrule
    \end{tabular}
    \caption{The training hyperparameters for each dataset.}
    \label{tab:hyperparam}
\end{table}
Specifically, we use the small variant of the Transformer's decoder \cite{vaswani2017attention} (L=6, H=768, A=6) as the backbone of CoAD. The disease head and symptom head are fully connected layers that map the hidden states of the decoder to the appropriate output space. In the fixed turn setting, the hidden state of the last symptom is used as input for the disease head. In the varied turn setting, the model terminates symptom inquiries by predicting the end token and proceeds with disease diagnosis immediately. The batch size and learning rate for each dataset are presented in Table \ref{tab:hyperparam}.
\section{Comparison of Fixed Turns}
\label{appdx:fix_turn}
\begin{figure*}
  \centering
  \begin{tabular}[b]{c}
    \includegraphics[width=.23\linewidth]{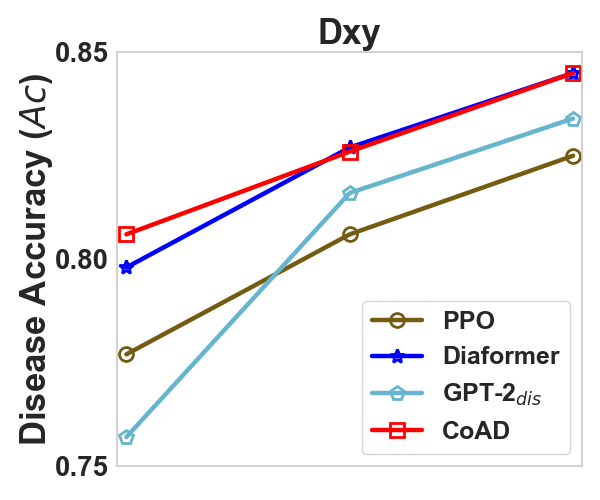}  
  \end{tabular} \hspace{-5mm} \vspace{-0.7mm}
  \begin{tabular}[b]{c}
    \includegraphics[width=.23\linewidth]{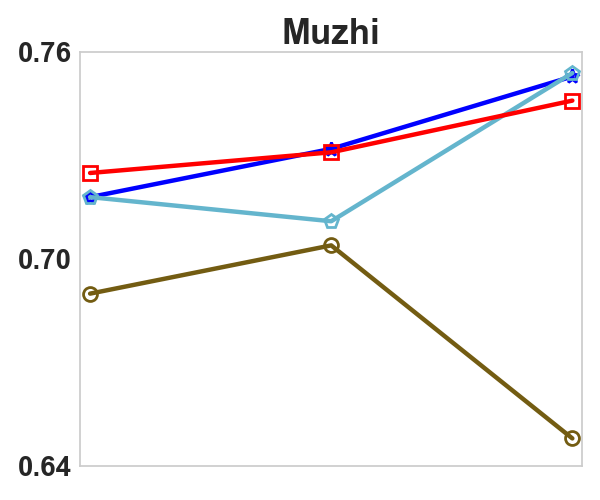}   
  \end{tabular} \hspace{-5mm} \vspace{-0.7mm}
  \begin{tabular}[b]{c}
    \includegraphics[width=.23\linewidth]{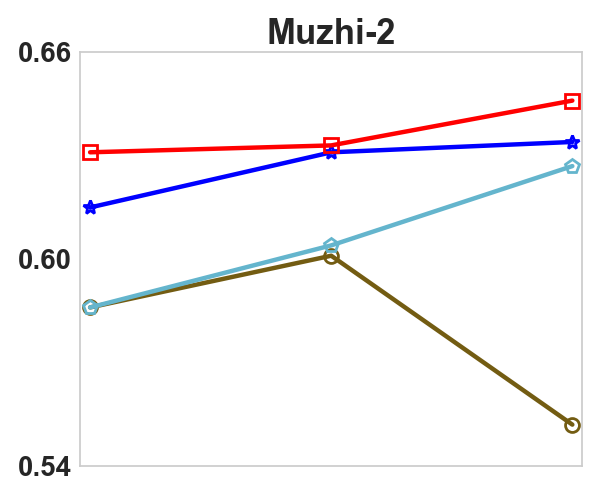}   
  \end{tabular} \hspace{-5mm} \vspace{-0.7mm}
  \begin{tabular}[b]{c}
    \includegraphics[width=.23\linewidth]{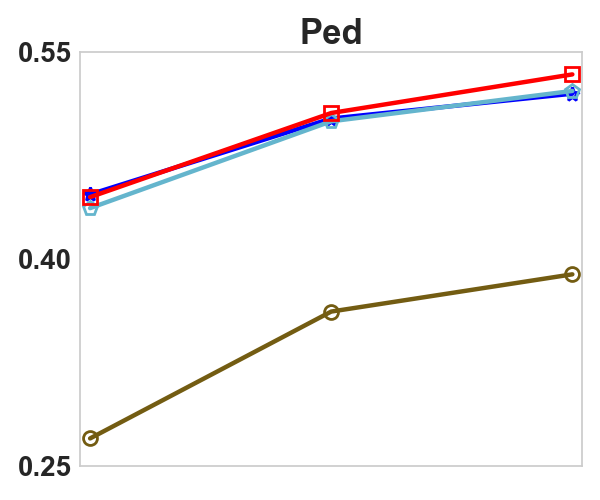}  
  \end{tabular} \hspace{-5mm} \vspace{-0.7mm}

  \begin{tabular}[b]{c}
    \includegraphics[width=.23\linewidth]{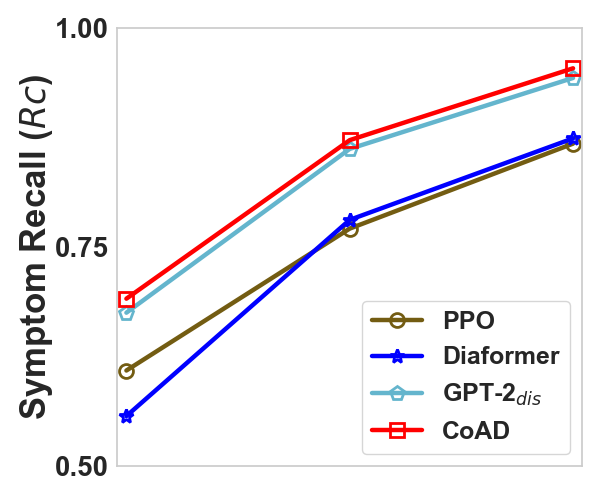}  
  \end{tabular} \hspace{-5mm} \vspace{-0.7mm}
  \begin{tabular}[b]{c}
    \includegraphics[width=.23\linewidth]{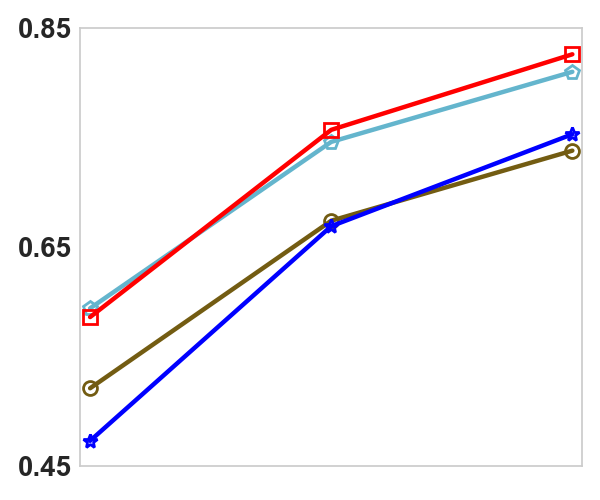}  
  \end{tabular} \hspace{-5mm} \vspace{-0.7mm}
  \begin{tabular}[b]{c}
    \includegraphics[width=.23\linewidth]{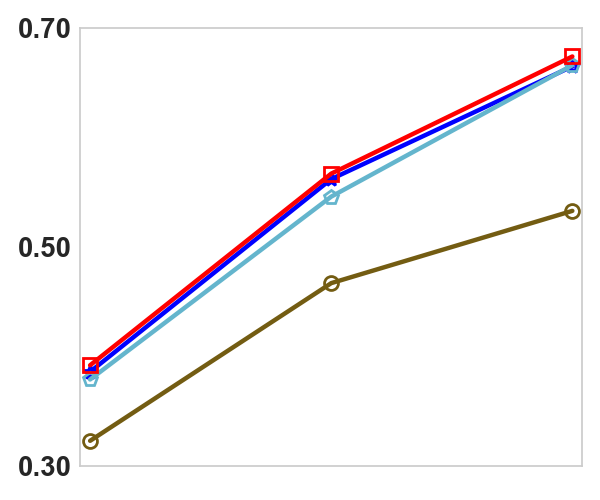}  
  \end{tabular} \hspace{-5mm} \vspace{-0.7mm}
  \begin{tabular}[b]{c}
    \includegraphics[width=.23\linewidth]{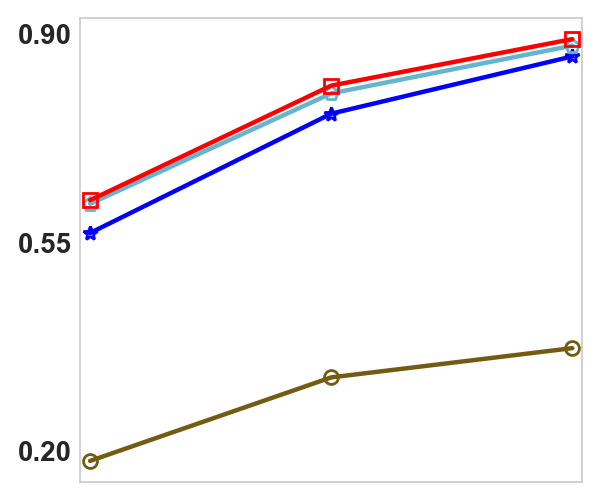}  
  \end{tabular} \hspace{-5mm} \vspace{-0.7mm} 

  \begin{tabular}[b]{c}
    \includegraphics[width=.23\linewidth]{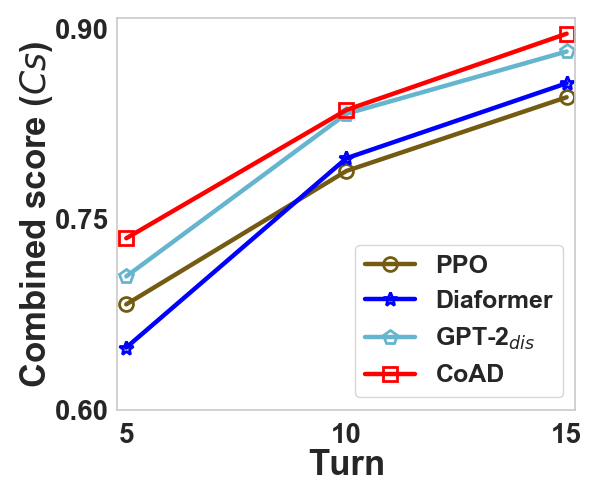}  
  \end{tabular} \hspace{-5mm}
  \begin{tabular}[b]{c}
    \includegraphics[width=.23\linewidth]{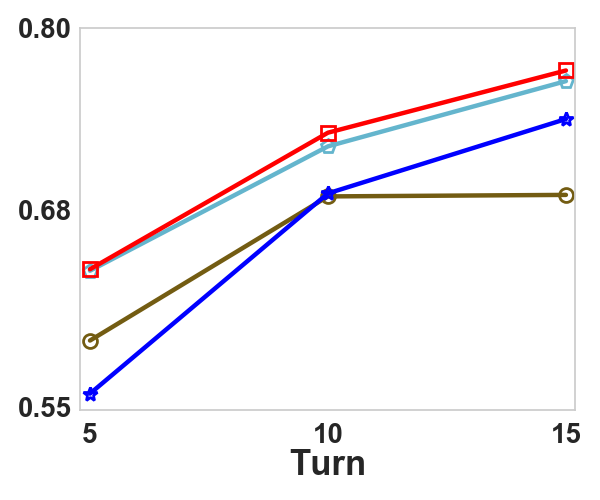}  
  \end{tabular} \hspace{-5mm}
  \begin{tabular}[b]{c}
    \includegraphics[width=.23\linewidth]{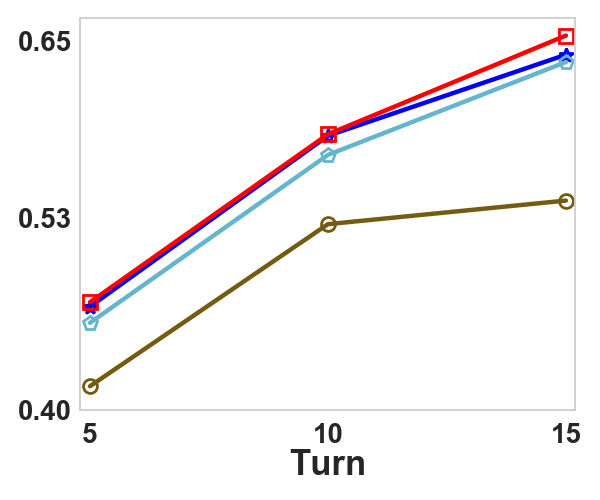}  
  \end{tabular} \hspace{-5mm}
  \begin{tabular}[b]{c}
    \includegraphics[width=.23\linewidth]{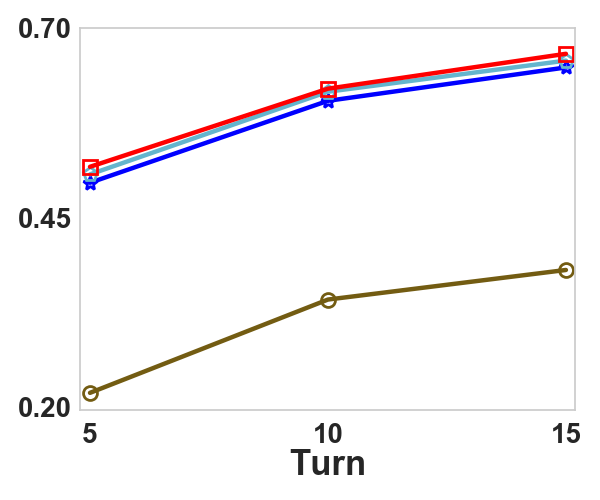}  
  \end{tabular} \hspace{-5mm}
  \caption{\label{fig:fix_turn} The average disease accuracy (Ac), implicit symptom recall (Rc), and combined score (Cs) of different models in the four datasets with 5, 10, and 15 fixed turns.}
\end{figure*}
\end{document}